*Article*

# Using Unmanned Aerial Systems (UAS) for Assessing and Monitoring Fall Hazard Prevention Systems in High-rise Building Projects

Yimeng Li[1], Behzad Esmaeili [2]*, Masoud Gheisari [3], Jana Kosecka [4], and Abbas Rashidi [5],

1. Ph.D. Student, Department of Computer Science, George Mason University, 4400 University Drive #MS 4A5, Fairfax, VA 22030, email: yli44@masonlive.gmu.edu
2. Assistant Professor, Sid and Reva Dewberry Department of Civil, Environmental, and Infrastructure Engineering; George Mason University; Fairfax, VA, 22030. E-mail: besmaeil@gmu.edu
3. Assistant Professor, Rinker School of Construction Management, University of Florida, Gainesville, FL, USA; email: masoud@ufl.edu
4. Professor, Department of Computer Science, George Mason University, 4400 University Drive #MS 4A5, Fairfax, VA 22030, email: kosecka@cs.gmu.edu
5. Assistant Professor, Department of Civil and Environmental Engineering, University of Utah, Salt Lake City, Utah, email: abbas.rashidi@utah.edu
* Correspondence: besmaeil@gmu.edu

**Abstract:** This study develops a framework for using unmanned aerial systems (UASs) to monitor fall hazard prevention systems near unprotected edges and openings in high-rise building projects. A three-step machine-learning-based framework was developed and tested to detect guardrail posts from the images captured by UAS. First, a guardrail detector was trained to localize the candidate locations of posts supporting the guardrail. Since images were used in this process collected from an actual jobsite, several false detections were identified. Therefore, additional constraints were introduced in the following steps to filter out false detections. Second, to properly detect floors and remove the detections that were not close to the floors, the research team applied a horizontal line detector to the image. Finally, since the guardrail posts are installed with approximately normal distribution between each post, the space between them was estimated and used to find the most likely distance between the two posts. The research team used various combinations of the developed approaches to monitor guardrail systems in the captured images from a high-rise building project. Comparing the precision and recall metrics indicated that the cascade classifier achieves better performance with floor detection and guardrail spacing estimation. The research outcomes illustrate that the proposed guardrail recognition system can improve the assessment of guardrails and facilitate the safety engineer's task of identifying fall hazards in high-rise building projects.

## 1. Introduction

Falls from height is one of the main causes of fatality in high-rise buildings. The risk of such fatalities can be significantly reduced by utilizing safety controls, such as guardrails and personal protective measures [1-4]. Although frequent and quality inspections of safety conditions on-site can be a leading indicator of safety performance [5-8], safety managers are facing several challenges that hinder their ability to conduct more frequent safety inspections in high-rise buildings. First, there are a limited number of safety managers in each company that may be located on construction sites across the county. Second, the large size and vertical construction of high-rise buildings make frequent inspections of such projects difficult. Increasing the number of safety inspections and observing hard-to-reach areas would significantly improve safety performance in high-rise construction projects.

One of the technologies that can be used to increase the number of safety inspections is Unmanned Aerial Systems (UASs). Using UASs to identify potential safety hazards can help safety managers take preventive measures to mitigate safety risks or provide warnings for workers who might be exposed to hazards. This study uses UASs and novel computational techniques to create an automated guardrail monitoring framework. The objective was to develop and test an image processing algorithm to identify missing guardrail posts from the video feed of the construction site collected via a UAS. This study developed and tested an automated algorithm for guardrail assessment from RGB (red-green-blue) images collected by a UAS from a high-rise building construction project. True-color values in RGB images were used to extract colors and filter images to identify targeted objects images. The research team first examined the technical development of this image processing algorithm and then implemented it in a high-rise construction project. Implementing and developing image processing algorithms to identify construction safety challenges is a novel approach to increase the frequency of inspections in high-rise buildings and subsequently reduce the likelihood of fatal accidents.

## 2. Research Background

### 2.1. Fall Protection and Guardrails

According to OSHA, construction workers working near an open edge 6 feet or more above lower levels should be protected by protection measures such as guardrails. Previous studies stated that relying on a single protection measure (e.g. only using fall arrest system as the last line of defense) cannot prevent falls, and potential fall hazards should be monitored continuously [9]. Because safety managers may not be available on a construction site to constantly monitor guardrail systems, safety researchers attempted to develop new automated methods to inspect construction sites. For example, some researchers developed algorithms to automatically monitor guardrail installations based on hazardous activities in a project schedule [10-11]. While this approach enables safety managers to monitor fall hazards, it requires having access to an as-built location-based measurement system of installed guardrails. Other researchers used computer vision techniques to detect construction guardrails [12-14]. While these studies contributed to the body of knowledge, limited studies have developed algorithms to assess fall protection systems drawing on UASs, mainly due to technical complexities in detecting such objects in a video.

### 2.2. Photo/Videogrammetry

Within the last two decades, advances in digital cameras, as well as emerging computational tools, have enabled practitioners and researchers to implement photo/videogrammetric techniques to process visual data (e.g., images, video) collected at construction jobsites and extract useful information by utilizing image processing, computer vision, and machine learning algorithms. Photo/videogrammetry is currently an active and ongoing research topic within the construction research community, and Table 1 summarizes some recent studies on applying such techniques for various construction applications.

**Table 1**. Summary of the recent studies on applying photo/videogrammetry for various applications in the construction engineering domain

| Objectives | Use cases | Citations |
| --- | --- | --- |
| 4D visualization and automated progress monitoring at construction jobsites | Contractors and Owners | [15-17] |
| Automated safety monitoring systems | Contractors and Owners | [18-21] |
| Damage and defect detection for a maintenance inspection and structural health monitoring | Architectures, Engineers, Contractors, Owners, Facility Managers | [22-24] |
| As-built documentation of buildings and other civil infrastructure scenes | Architectures, Engineers, Contractors, Owners, Facility Managers | [25-27] |
| Volumetric surveying, quality control, and tolerance measurement | Contractors, Fabricators | [28-29] |
| Generating 3D building thermal profiles for energy modeling and analysis | Engineers, Contractors, Owners, Facility Managers | [30-31] |

The primary advantage of using photo/videogrammetric approaches is that unlike using active methods such as wearable sensors, there is no need to attach these visual sensors to the objects of interest. Besides, the data collection procedure is inexpensive and straightforward as off-the-shelf cameras are ubiquitous nowadays. On the other hand, limited fields of view, sensitivity to lighting conditions, and the existence of obstacles and barriers are the significant limitations of implementing such techniques [15].

Automated object detection via images and videos is a significant component of any photo/videogrammetric system. Subtracting background using various methods such as Gaussian Mixture Model (GMM) and Bayesian-based models and then implementing classifiers such as Bayes or neural networks [15] are standard procedures for object detection. More advanced feature-based recognition methods, such as Histogram of Oriented Gradients (HOG), have been recently employed by researchers to detect various objects at construction jobsites [32]. Tables 2 and 3 summarize recent studies on computer-vision-based methods for detecting two significant classes of construction resources: workers and heavy equipment.

**Table 2.** An overview of recent studies (2015 and newer) for implementing computer vision-based methods for detecting construction equipment [15]

| Objectives | Methods used | Citations |
|---|---|---|
| Crowdsourcing video-based activity analysis | HOG object detector and tracking. User annotations on activities | [32] |
| Safety assessment through crowdedness and proximity estimation | GMM background subtraction and Kalman filter tracking | [33] |
| Estimate production cycles of loading activities | GMM Background subtraction and kernel covariance tracking | [34] |
| Safety assessment and warning | GMM background subtraction and HOG for detection and Kalman filter tracking | [35] |
| Semantic annotation of construction videos | HOG object detection and frame similarity measurement | [36] |
| Estimate production cycles of loading activities in tunneling | Region-based fully convolutional networks | [37] |
| Estimate production cycles of loading activities | Tracking-Learning-Detection (TLD) algorithm | [38] |
| Proximity monitoring between mobile resources | CNN object recognition | [39] |
| Estimate production cycles of hauling activities | License plate detection and recognition (LPDR), and deep convolutional network | [40] |
| Estimate production cycles of hauling activities | HMM, atomic action recognition, deep learning-based detection, and tracking | [41] |
| Estimate productivity and cycle time of earthmoving operations | Faster Recursive CNN (R-CNN) and TLD | [42] |

**Table 3**. An overview of recent studies (2015 and newer) for implementing computer vision-based methods for detecting construction workers [15]

| Objectives | Methods used | Citations |
|---|---|---|
| Crowdsourcing construction activity analysis | HOG and HOC features | [32] |
| Worker action recognition using the dense trajectories method | HOG, HoF, and Motion Boundary Histogram (MBH) | [43] |
| Monitor and analysis of installing reinforcement activities in construction | RGB, optical flow, and gray stream | [44] |
| Biomechanical analysis and/or ergonomic posture assessment in modular construction | Joint and body part detection, 3D body model generation, and joint angle calculation | [45] |

Other than detecting and tracking construction resources, photo/videogrammetric techniques have been recently utilized for automatically detecting safety-related objects and devices at construction jobsites, including Personal Protective Equipment or PPE [15,18-20], fall protection devices [18,19], and guardrails [12]. The ultimate goal of these studies is to automate efficient safety monitoring and management systems. The proposed method in this study also falls into this category by automatically detecting guardrail posts via images and videos collected by UAS.

*2.3. Unmanned Aerial Systems (UASs)*

Unmanned Aerial Systems have been used for various construction applications: progress monitoring [46-48], earthmoving assessments [33, 49-52], building inspection [53-56], material handling [57,58], and safety management [59-63].

UASs can improve safety performance on construction jobsites. They can carry data collection sensors (e.g., video cameras, heat or motion detectors, laser scanners) to transfer real-time or processed data to construction safety managers [64,65]. UASs can fly and move fast to unsafe or hard-to-access locations on job sites. UASs can perform some tasks quicker, safer, and at a lower cost than some manned vehicles [66]. Irizarry et al. [67] and Gheisari et al. [68] used a quadcopter equipped with a video camera sensor to provide real-time video feed of a construction project to safety managers for inspection purposes.

In one study, Roberts et al. [63] used a UAS system equipped with an object detector sensor to identify and reduce safety hazards associated with cranes. Gheisari et al. [59] used Point Cloud Data (PCD) generated from UAS-acquired videos to identify potential fall hazards on the site. UAS-acquired visuals or the generated PCDs were used to identify unsafe conditions or hazards in several other studies [60,62,69]. In a recent study, Martinez et al. [61] developed iSafeUAS, a UAS platform specifically designed and developed for construction safety inspection purposes. iSafeUAS uses the super optical zoom capability of an advanced high definition camera and a parachute recovery system to identify better and inspect on-site safety hazards while significantly reducing the on-site UAS fall risks [61]. Gheisari and Esmaeili [65] recently conducted a national survey study with safety managers about UASs for construction safety monitoring. That study identified monitoring the work near an unprotected edge/opening using UASs as one of the top application areas indicated by safety managers [65]. Therefore, this study aims to analyze videos and images collected from UASs using machine learning and computer vision techniques to assess guardrails protecting openings and edges.

While replacing human eyes with cameras mounted on UASs is appealing, the current practices for processing such data are still semi-automatic and require some manual steps and conducting measurements by end-users. To tackle this issue, and in recent years, several customized image processing algorithms have been introduced by researchers for various applications within the architecture, engineering, and construction domains [17]. Akbar et al. [70] proposed an automated structural health monitoring (SHM) system for high-rise buildings and skyscrapers based on coupling UASs and image processing techniques. Gopalakrishnan et al. [71] suggested an automated crack damage detection system based on processing images collected by UASs.

Automated progress monitoring at construction jobsites is another potential application of integrated image-based UASs: Asadi et al. [72] proposed an integrated Unmanned Aerial/Ground Vehicles (UAV/UGV) system for collecting useful visual data from construction jobsites. Ibrahim and Golparvar-Fard evaluated a 4D BIM-based optimal flight planning for construction monitoring purposes [73]. Roberts et al. [63] implemented image processing algorithms to detect and classify cranes from UAS-based imagery data. Processing visual data collected by UASs could also automate asset management and maintenance procedures [74].

## 3. Research Objectives

In this study, the authors investigated the possible application of UASs for assessing guardrail systems close to unprotected edges and openings. The specific research objectives of this study are twofold: (1) Developing a hazard identification framework based on using UASs and focusing on guardrails and (2) Evaluating the feasibility of the proposed framework by conducting a case study in a real-world high-rise construction jobsite. Using UASs, the research team developed and implemented an automated guardrail assessment framework that could facilitate inspection tasks of safety personnel at larger size construction jobsies.

Currently, there are several existing studies in the literature discussing the potential use of UASs and mounted cameras for automating safety inspections at construction jobsites. Such studies mainly focus on hardware and data collection requirements and path and trajectory planning rather than developing necessary automated algorithms for detecting and analyzing hazardous situations [60, 61]. As a result, this research focuses on automatically detecting guardrail posts as one of the essential safety features at vertical construction jobsites. Implementing computer vision-based methods for identifying safety hazards is an ongoing and demanding research area. Compared with other computer vision-based methods, our proposed method contains the following advantages: The majority of the existing methods in the literature are designed and implemented at horizontal construction jobsites. Our approach has been implemented and tested in a high-rise building with challenging data collection and analysis settings of vertical construction. Existing studies are often limited to only detecting objects of interest. Our proposed method can handle floor detection and, more importantly, space estimation to detect the guardrail. This proposed system could eventually be used as an automated safety compliance checking and monitoring system of guardrails.

## 4. Research Framework

A three-step framework is developed to identify guardrail posts from the images captured by UAS from high-rise buildings (see Figure 1): (1) guardrail detection, (2) floor detection, and (3) spacing estimation.

During the first step, possible locations of the posts that support guardrails are identified by training the guardrail detection algorithm. One possible issue with this process is the higher chance of producing false detection results due to the fact the images taken from a diverse construction jobsite are not very consistent. To tackle this issue, a number of additional constraints are implemented within the next two steps to filter out the false detection cases. As the next step, and since the guardrails are installed on different floors close to edges, a horizontal line detector is introduced to locate floors and remove the detection cases that are not in the vicinity of the floors. As the third step, and considering the fact that the distances between neighboring guardrail posts are pretty consistent, the space between them is approximated to detect the most probable combination of their locations. These steps will be explained in more detail in the following subsections.

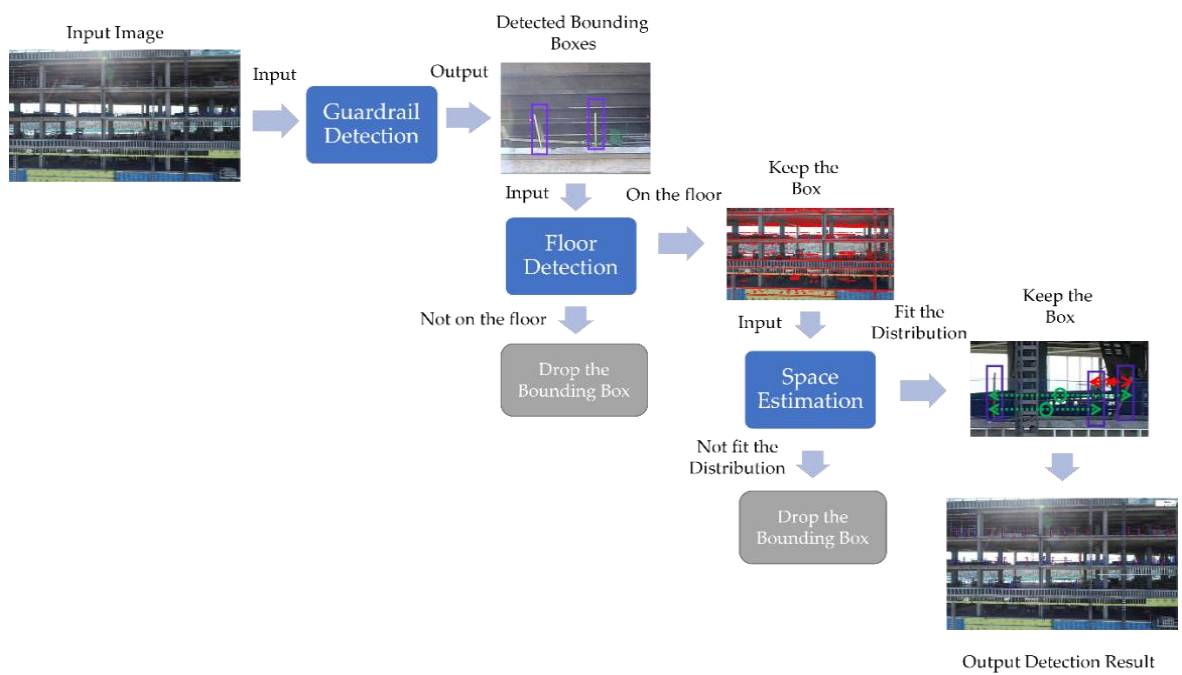

**Figure 1**: The overall framework for detecting guardrails using UAS-based visual data

### *4.1. Step#1: Guardrail Detection*

We use the sliding-window approach to search for guardrail posts in the image. A sliding window is a rectangular region of fixed width and height in computer vision. The presence of the object of interest is evaluated at a dense set of locations in the image. We apply an object classifier for each of these window regions to determine if the window has an object of interest, in this case, a guardrail assembly (i.e., posts and rails). The sliding-window method has been successfully used in face recognition [75] and pedestrian detection [76]. In this method, a window is slid through the whole image from the upper left window of the image to collect candidate locations likely to contain a guardrail assembly. Fixed-size windows are used, assuming knowledge of the size of the guardrail's projection in the images obtained from the 3D reconstruction of building meta-data. This assumption helped to improve the efficiency of the sliding window approach. A histogram of oriented gradients (HOG) features is extracted from each window. This HOG feature is one of the most valuable features for object detection introduced in pedestrian detection applications [76]. Given some ground truth examples of guardrail posts, a binary classifier is trained to process if the window contains a guardrail assembly or not. The classifier takes its HOG feature for each window and predicts a score. This window is considered a positive detection if the score is above a specific threshold.

The research team has considered different classification methods for the analysis, including cascade classifier, support vector machine (SVM) classifier, artificial neural networks, and deep convolutional neural networks. Since neural networks do not consider spatial dependencies inherent in image data are not suitable for the analysis. While deep convolutional neural networks can learn the features for the classification tasks, they require a significant number of training examples and are better suited for multi-class classification tasks [86, 87]. Considering the access to a small amount of hand-labeled ground-truth training data in this study and guardrail detection being a binary classification task, deep convolutional neural networks are not suitable. They are likely to overfit n the training data. In contracts, Cascade classifiers and linear SVM are two robust classifiers that are a better fit for binary classification tasks and were used in this study [75, 76]. These methods have proven to be generalizable to various image data types and require a moderate number of training examples. Cascade classifier can also learn the features simultaneously and has been successfully applied for face detection in several commercial settings. Furthermore, linear SVM (support vector machine) was used in this study instead of kernel SVM because (1) the HOG feature extracted from images are linear features [89], and (2) after comparing the linear SVM performance against radial basis function (RBF)-kernel SVM, Linear SVM achieved better results and was more efficient [90].

The following two classification algorithms are suggested in this phase, 1) cascade and 2) linear SVM. Although, both these algorithms are efficient in making decisions and easy to implement, in order to minimize the number of overlapping detections, the research team adopted the NMS (non-maximum suppression) algorithm [77].

The IOU (intersection over union) metric is used to measure the performance of algorithms. IOU is the ratio of the overlap between two areas of the ground truth bounding box A and predicted bounding box B over their union and can be calculated using equation 1 (Figure 2) [77].

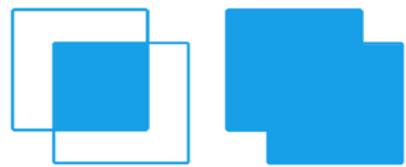

**Figure 2**: Area of overlap (left) vs. area of union (right)

$$IOU = \frac{Area\ of\ Overlap}{Area\ of\ Union} \qquad (1)$$

*(a) Cascade Classifier*: a cascade classifier that has been used for efficient face detection in previous studies [75], is explicitly trained for guardrail detection. Guardrail labels are prepared to train the captured data during the training stage. Using the labels, a classifier is trained, using the cascade of classifiers, to detect the presence of guardrail assemblies in image regions. The cascade employs more complex binary classifiers at each stage. for example, one region is split into more cells when computing the HOG feature. As a result, the algorithm can quickly reject regions that do not contain the target. In the case of not finding the desired object at any cascade stage, the detector rejects the window region and terminates the process. Many object candidate regions have low confidence at the early stages, making the cascade classifier efficient. An off-the-shelf Cascade Object Detector from MATLAB [78] can be adopted for training purposes.

*(b) Linear SVM Classifier*: a linear SVM algorithm is used to classify the candidate windows into guardrail and background. The SVM model learns a decision boundary in the HOG feature space to perform classification. To find the best set of hyperparameters, we adopt a cross-validation process with the grid search algorithm to find the best set of hyperparameters is adopted. A grid search algorithm is applied for selecting the parameters. Grid search algorithm refers to the technique for exhaustive search of optimal parameters combination. For two parameters where each can take n values, the results are saved into a n x n grid. The optimal combination of the two parameters is obtained by a 2D grid search over combinations of the values of parameters C and $\gamma$., C is the regularization term. The strength of the regularization is inversely proportional to C. $\gamma$ is the kernel coefficient, defining the influence of one training example [79]. Each block in the grid represents a pair of parameter values. For each pair of parameter values, we train a linear SVM model and evaluate its performance on the validation set using the comprehensive search results. Finally, the best-performing SVM model is selected. The *libsvm* library [79], a library for Support Vector Machines, can train the SVM classifier.

*4.2. Step#2: Floor Detection*

Identifying the horizontal segments of the building floor begins by detecting the vanishing points and the parallel lines associated with them. Since there are many parallel horizontal lines in the used building images, horizontal segments are identified by determining the largest parallel line group with a vanishing point, and floors are detected by a large number of detected line segments. Since images from actual construction projects will be used in this study, the parts can be inconsistent, leading to separate parts belonging to the same horizontal lines. To avoid dealing with all the small line segments, the small pieces can be clustered into long ones by merging elements with similar intercepts. Then by assessing the coverage of each clustered line segment on the x-axis, one can consider the maximum ones as the detected floor. For example, the top ten lines were picked as the detected floors for an image of a building with three floors as shown in Figure 1. These detected floors are then used to filter out false-positive detections. The number of false detections can be reduced using a threshold that considers the distance from the bottom of a detected bounding box to the closest detected floor. The details of the vanishing point detection algorithm can be found in [80].

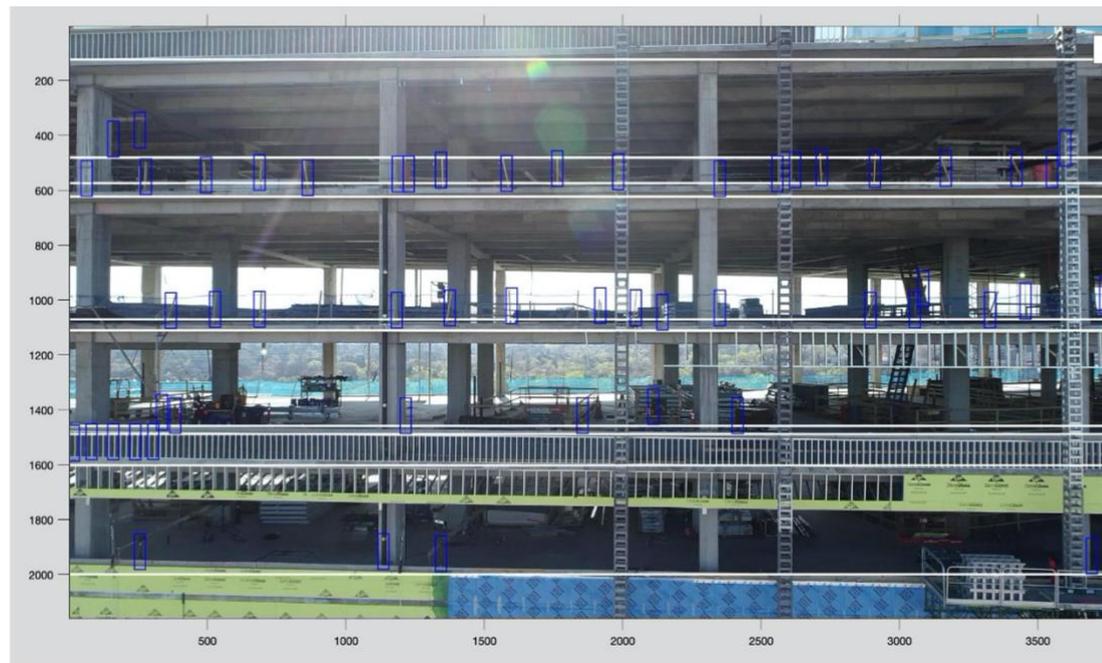

*Figure 1 Detected Floors*

*4.3. Step#3: Space estimation*

First, we assess the space between neighboring guardrail posts in the training images and try to fit a spatial distribution model through Expectation-Maximization (EM) algorithm. The aim of this space estimation is to detect missing guardrail posts, and it does not intend to check whether guardrail elements are in exact compliance with standards. After carefully observing the data, it was noticed that the Gaussian Mixture Model (GMM) is an accurate estimation of the space between neighboring guardrail posts in the training set. GMM model means the data distribution follows a mixture of multiple Gaussian distributions. A GMM model's parameters include the number of Gaussian components and the mean and variance of each Gaussian component. The EM algorithm is a technique to decide GMM's parameters. To check whether GMM provides a reasonable estimate, a plot of the spatial distribution among the training data can be visually observed [81]. The space between guardrail posts can be computed as the center distance between neighboring bounding box annotations while ignoring guardrail posts without neighbors. The output of the first step is a general space rule. Then we apply dynamic programming to find the best combination of detections with the most plausible spacing. Detections disobeying the general space rule are removed. Dynamic programming helps us efficiently search through the combinatorial space.

**5. Case Study: Implementing Guardrail Detection Algorithm In A Real-World Project**

An under-construction 16-story high-rise commercial building project was selected as the case study site for this project (Figure 3). This specific project was selected because (1) it was a high-rise construction project with a significant amount of safety guardrails being used, (2) there was a small amount of vegetation (e.g., trees) around the building and vegetation will block the UAS view when capturing the video and (3) the building was not being on a Federal Aviation Administration (FAA) restricted airspace zone. The process for monitoring guardrails started with traversing the scenes using the videos captured by a UAS. To minimize occlusions, the videos were captured from various directions and angles. The research team deployed a DJI Phantom 4 PRO Quadcopter to record a series of videos of the guardrails from the high-rise building project, as shown in Figure 4. The UAS platform used provided a high-quality camera that shoots 4K videos at 60fps and photos at 20 megapixels, and its battery provided roughly 30 minutes of flight time under optimal conditions. This approach has been successfully used in previous studies in the construction industry [83,84]. A combination of manual and semi-autonomous flight capabilities (e.g., course lock, waypoints, and point-of-interest) were used during the data collection. To comply with the FAA's Small UAS Rule (14 CFR part 107), there was no fly directly over workers [82], so the flight paths were designed to ensure no workers or live traffic under the flight path.

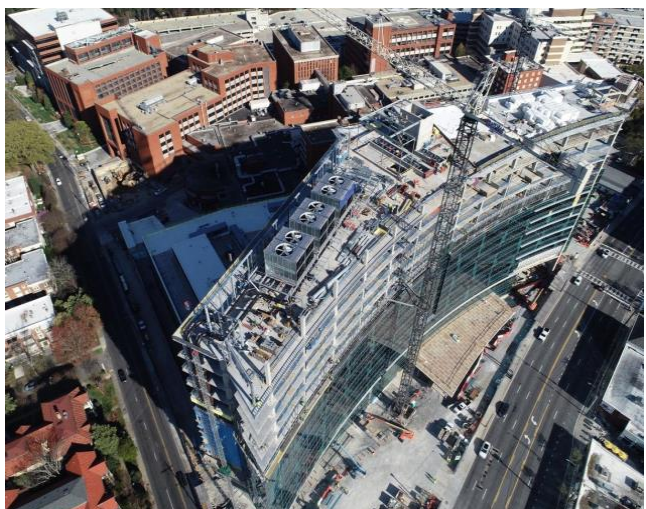

**Figure 3**: Case study project site

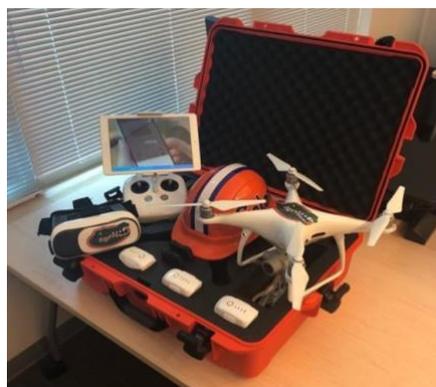

**Figure 4**: Phantom 4 PRO Quadcopter UAS

The research team conducted multiple flights while changing flight conditions and settings (e.g., manual and autonomous navigation, proximity to buildings, variations in speed, and image percentage overlap) to cover almost all the building facades. Since the focus of the study is on fall protection, the relevant visual information related to fall hazards and guardrails was selected for further analysis.

From a 5 minute video taken by the UAS circling around an under-construction building, 68 keyframes were extracted. The whole keyframe extraction process is automatic. We sample the keyframe from every 100 frames. The original writing is misleading. The video is 5 minutes 28 seconds, and the fps is 24, meaning 24 frames per second. For the first 20 seconds and the last 20 seconds of the video, the UAS is tuning its pose, so we skip these frames. So the final number of keyframes is ((5min*60s+28s)-40frames)*24fps/100 = 68. To minimize the similarity between the selected images, the research team used a sampling rate of about 100 frames per second. The research team split images into training and testing data to develop and test the algorithm. Fifty images are used for training and the remaining 18 images for testing. Cross-validation is done on the training set, where one portion of the training set is used as the validation set. All the guardrail assemblies were manually labeled in all the images. Typically there are between 18 to 30 guardrail assemblies in each image. As far as the training data was concerned, 1158 guardrail posts were labeled. In the testing data, the research team labeled 416 guardrail posts.

*5.1. Evaluating Performance in Detecting Guardrail Posts*

As stated earlier, the research team trained a cascade classifier and a linear SVM on the training data. To reduce the number of false detections, 10 pixels distance was selected as the threshold according to image resolutions. using this threshold, the detections with more than ten pixels above the detected floor were classified as the false positives. The testing data was later used to evaluate performance by calculating "Precision" and "Recall" metrics. The equations that were used to compute the metrics are:

$$Precision = \frac{t_p}{t_p + f_p} \quad (2)$$

$$Recall = \frac{t_p}{t_p + f_n} \quad (3)$$

*tp*, *fp*, and *fn* is the short form of true-positive, false-positive and false negatives. True positive is the number of detected guard rail boxes whose intersection with ground truth is above the threshold. False-positive is the number of detected guard rail boxes without an intersected labeled guard rail. False-negative refers to the labeled ground truth guard rails missed by the detector (i.e., have no intersection with detected boxes) [77]. It is worth mentioning that the linear SVM performed better (10 percent higher recall) than the cascade classifier (Figures 5 and 6), which means that the linear SVM detected the guardrail posts more frequently in the image (Table 4). However, the linear SVM'sprecision is significantly lower. Comparatively, the cascade classifier achieved a balance between precision and recall.

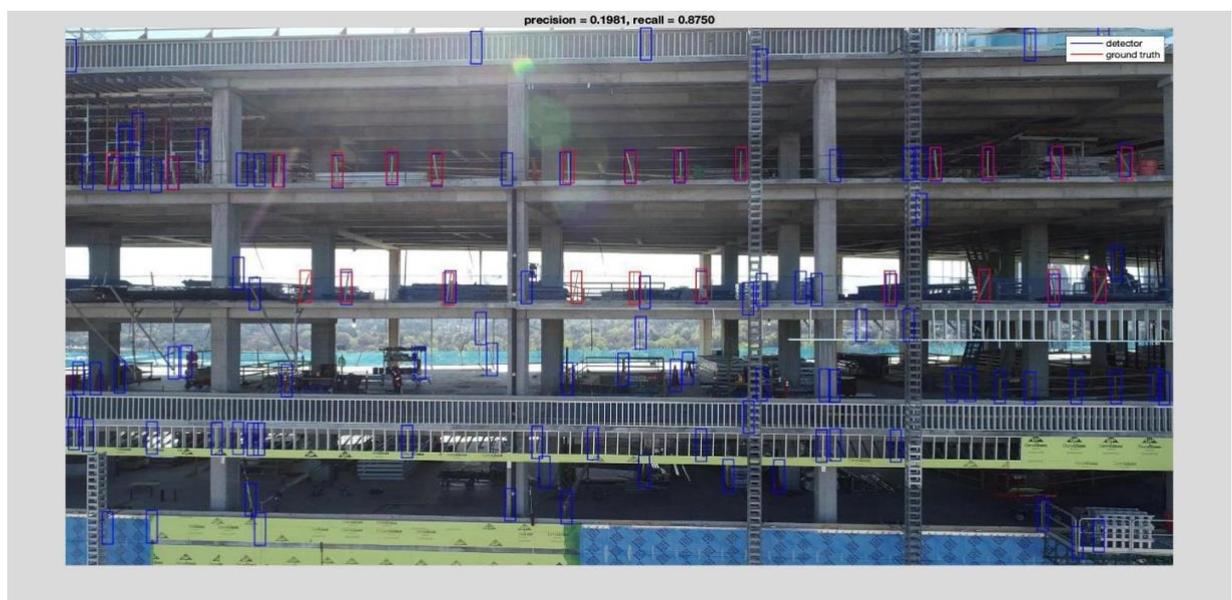

**Figure 5**: Visualizations of the guardrail detections using linear SVM

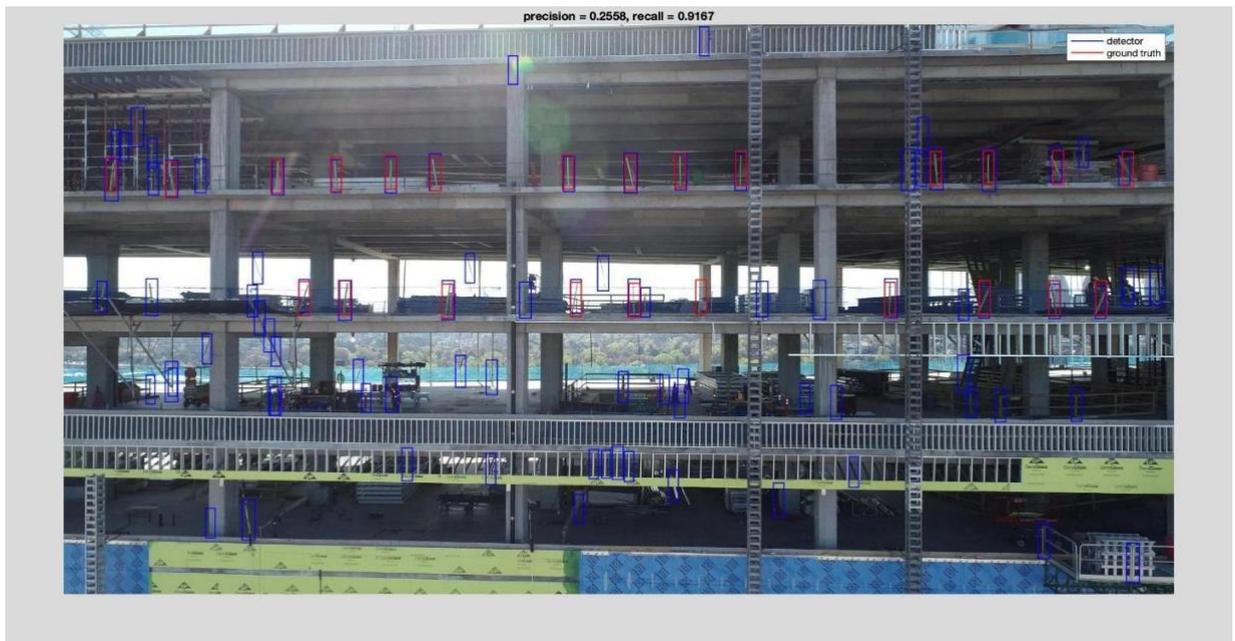

**Figure 6**: Visualizations of the guardrail detections using the cascade classifier.

**Table 4**: Results of evaluation for trained guardrail detectors

| Metric | Precision | Recall |
| --- | --- | --- |
| Cascade Classifier | 0.1510 | 0.7077 |
| Linear SVM | 0.0438 | 0.8062 |

*5.2. Evaluating the Guardrail and Floor Detection*

As mentioned earlier, first, the floors were detected on the test images. Then, the floor detection filtering was applied to the detected windows. The results shows that many false-positive detections were removed after integrating the floor detection. Moreover, the precision of both cascade and linear SVM classifiers almost doubled. The effectiveness of the floor detection filtering on the linear SVM result are shown in Table 5 and Figure 7.

Table 5: Results of evaluation after applying the floor detection step

| Metric | Precision | Recall |
| --- | --- | --- |
| Cascade Classifier and Floor Detection | 0.2531 | 0.7015 |
| Linear SVM and Floor Detection | 0.0974 | 0.7908 |

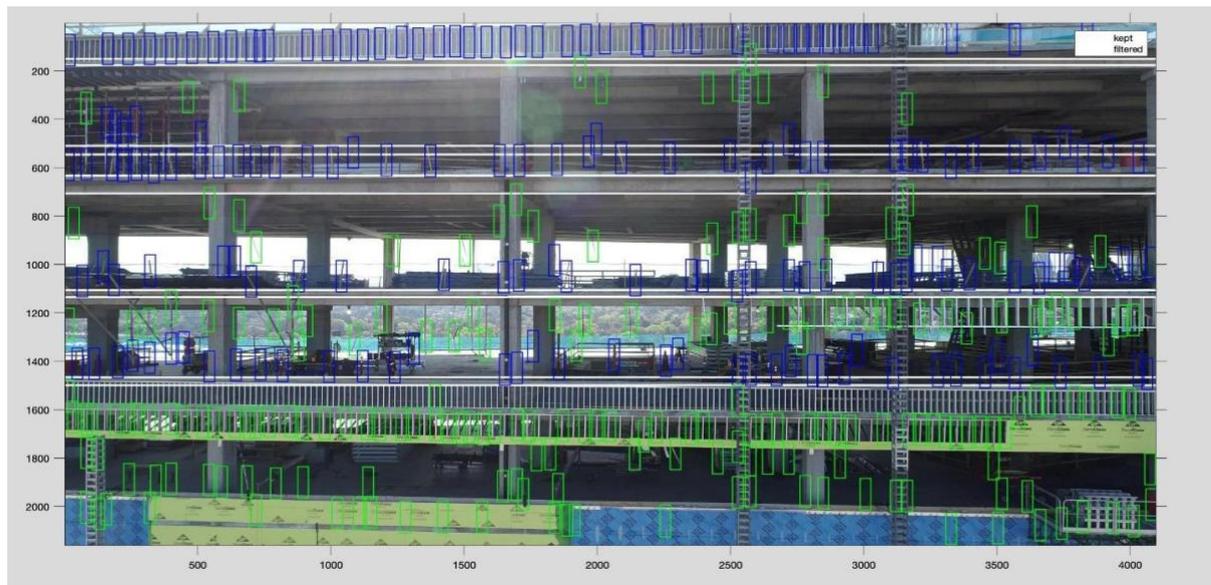

**Figure 7**: Visualization of the filtered false positive windows (denoted as green bounding boxes). White lines are the detected floors.

*5.3. Evaluating the Guardrail Detection, Floor Detection, and Space Estimation*

This study used space estimation as the final post-processing technique to achieve the best combination of guardrail detections. Since the images were taken from different perspectives, the spaces need to be normalized. We normalized the space between guard rails by dividing the space between two guardrail posts with the median space of all the guardrail pairs on the same floor. In more detail, all the guardrail bounding boxes on the same floor in each image were collected and the guardrail spaces on the same floor were normalized. After space normalization, we collected the guardrail space, estimated the distribution, and summarized it in a histogram. At the expectation step of the EM algorithm, posterior probabilities of component memberships were computed and later were used as weights to estimate the component means, covariance matrices, and mixing proportions by applying maximum likelihood. The results estimated the number of components to be three, meaning three normal distributions were found (Figure 8). Also, a "space-ubiquity" table was built so one could use a space value to predict the corresponding space value of the training set.

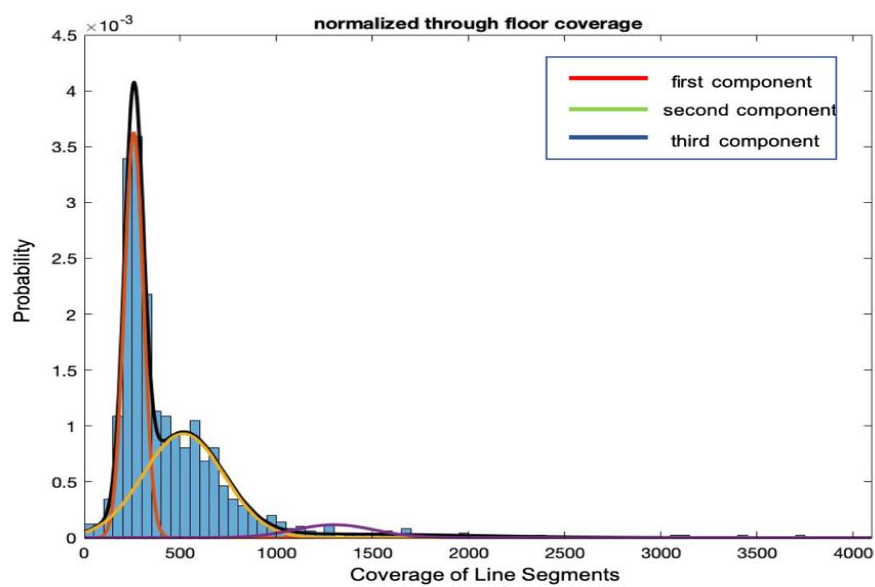

**Figure 8**: The space distribution among the training data are visualized here. The approximations for the 3 components are shown in form of the colorful bell curves.

Given the group of bounding box detections during the testing stage, the goal was to find their combination with the maximum ubiquity. This goal was achieved by first computing the ubiquity value between every pair of detections through the space-ubiquity table and then finding the maximum combination using dynamic programming (DP). Notably, the ubiquity value at each space is reduced by a certain threshold, which makes it negative for some spaces. The cascade classifiers yielded an overall recall of 62 percent and a precision of over 35 percent (Table 6). On the other hand, while using linear SVM detection reduced the precision by 10 percent, it increased recall by around two percent.

**Table 6**: Results of evaluation after applying space estimation

| Metric | Cascade Classifier + Floor Detection + Space Estimation | Linear SVM + Floor Detection + Space Estimation |
|---|---|---|
| Precision | 0.3666 | 0.2680 |
| Recall | 0.6215 | 0.6400 |

Moreover, the performance was not superior on a small set of testing images. For example, using a cascade classifier on some of the testing images yielded over 80 percent recall and 50 percent precision. A summary of evaluation results is shown in Table 7, and visual representations of the final results on two images are shown in Figure 9.

**Table 7**: Summary of evaluation for all proposed approaches

| Metric | Precision | Recall |
|---|---|---|
| Cascade Classifier | 0.1510 | 0.7077 |
| Linear SVM | 0.0438 | 0.8062 |
| Cascade Classifier and Floor Detection | 0.2531 | 0.7015 |
| Linear SVM and Floor Detection | 0.0974 | 0.7908 |
| Cascade Classifier and Floor Detection and Space Estimation | 0.3666 | 0.6215 |
| Linear SVM and Floor Detection and Space Estimation | 0.2680 | 0.6400 |

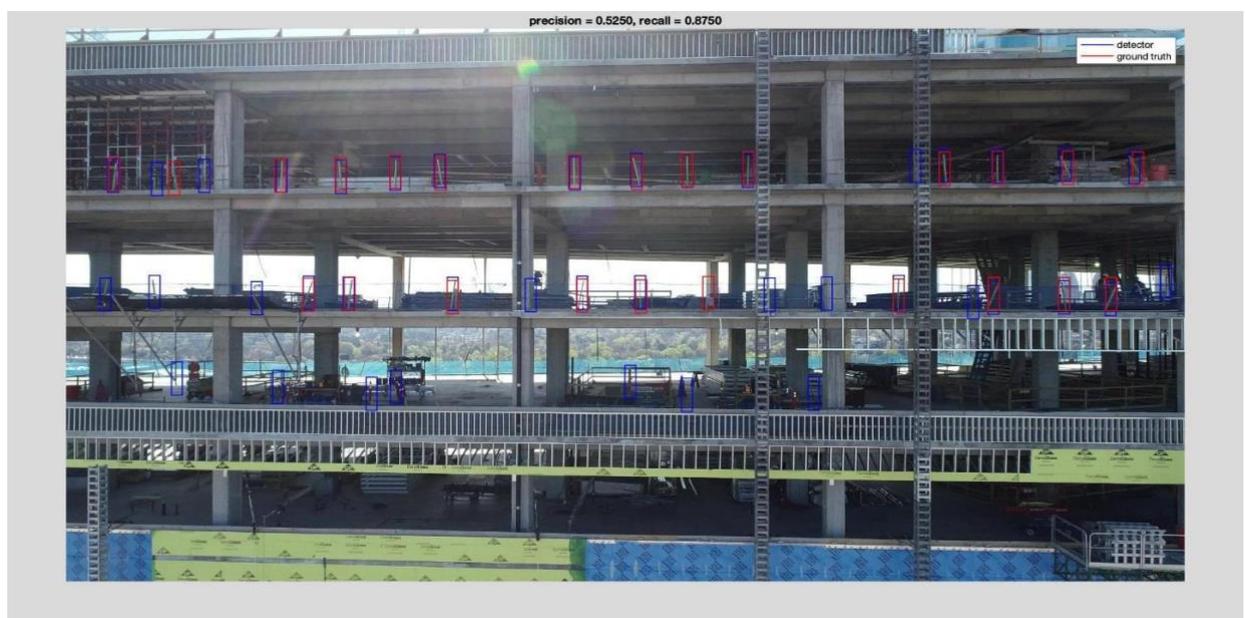

(a)

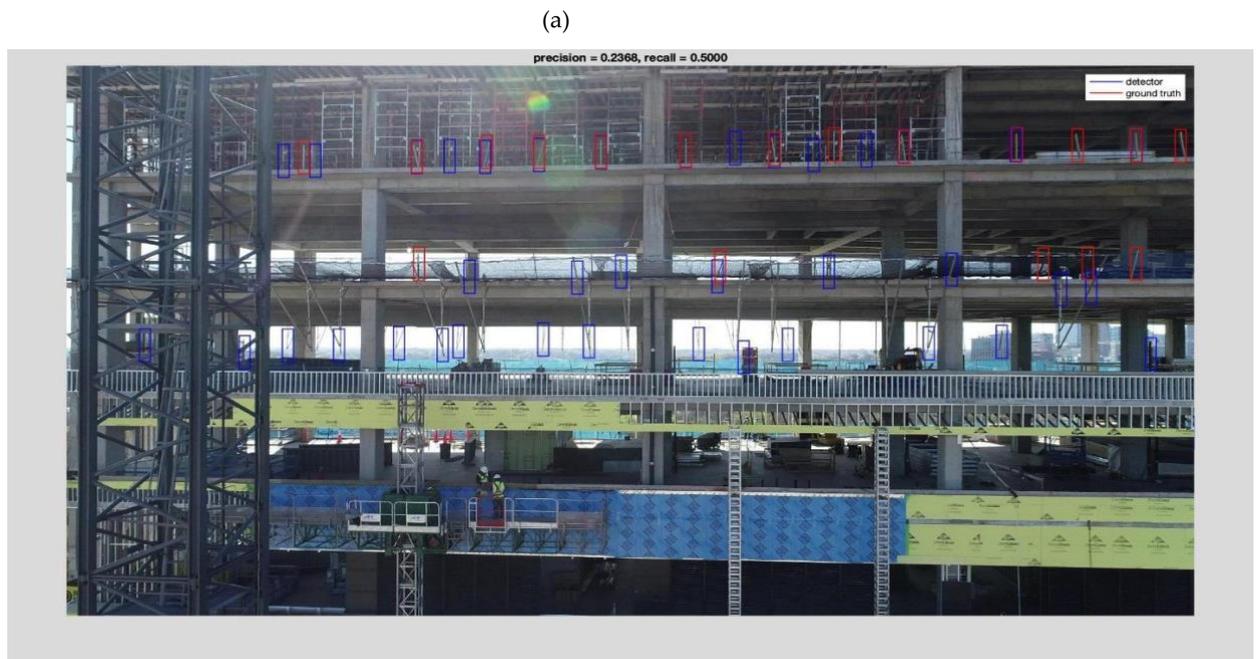

(b)

**Figure 9**: Final Detection Results on Two Images: (a) higher precision and recall and (b) lower precision and recall. The detected windows are specified with the blue bounding boxes, and the ground truth labels are illustrated with the red ones.

## 6. Conclusions

Increasing the frequency of safety inspections and monitoring hard-to-reach or inaccessible regions of construction job sites can help improve safety performance. This study developed and tested an image processing algorithm for guardrail posts detection in a real-world high-rise construction project. The main contributions of the study are summarized here:

- A three-step machine-learning-based pipeline (i.e., guardrail detection, floor detection, and space estimation) was proposed to detect guardrail posts in the images captured by UAS. In order to improve the overall performance of the proposed framework, the floor detection and guardrail spacing estimation were applied to filter false positive detections.
- Cascade classifier and linear SVM classifier and various combinations of these techniques were used to detect different elements of a guardrail system.
- The performance measures show that using the cascade classifier combined with floor detection and guardrail spacing estimation provides the best performance.
- One of the main contributions of this study is developing a framework to improve the accuracy of detecting guardrail posts from RGB images using a stepped approach.
- This study could further facilitate the safety engineer's task of identifying fall hazards in high-rise construction projects using data collected by UAS.

There are some limitations worth mentioning. First, the framework only used three steps to refine the guardrail detection, and future studies should be conducted to include more advanced steps to generalize the proposed framework. Second, the framework is tested for one case study. The framework needs to be tested in more case studies by collecting data from different high-rise buildings. Collecting more data would also make it possible to use convolutional neural networks for data analysis. Third, since the framework developed in this study only detects the posts of the guardrails and not the railings, there is a need for further visual inspection by safety managers. Furthermore, this study helps safety managers to determine whether guardrail systems are properly located and do not detect fall hazards that are not protected by guardrails. The framework should be expanded in future studies to address these limitations. Despite these limitations, this study develops a novel framework to detect guardrails using images collected from UASs automatically.

The next step of this study includes further development of the guardrail detection algorithm to not only include the posts but also consider the railings as another key element of the guardrail systems. In addition, guardrail systems for the stairs that might not necessarily follow the horizontal placements of the regular guardrail posts and might have more complex placements should also be integrated with the further development of this guardrail detection algorithm.

It should also be noted that the integration of UASs for construction safety purposes might have some other challenges that require further attention. First, introducing UASs to the construction sites might raise novel occupational safety and health issues (e.g., workers distraction, fall over people, struck-by hazards, psychological impacts) that require further investigation for their successful integration in the construction domain. Moreover, it is worth noting that UASs are currently capable of only playing the role of visual inspectors on the job sites. For a comprehensive safety monitoring system, UAS should also be integrated with other types of technologies and also work in collaboration with workers and safety managers on the ground. UASs should play an efficient role in the human-UAS-motivated safety culture to ensure complete compliance on the construction sites.

**Author Contributions:** All authors contributed to the idea and concept of this study. Software development, Y.L.; validation, Y.L. and J.K.; formal evaluation and analysis: Y.L. and J.K.; writing—original draft preparation, Y.L., B.E., M.G., J.K, and A.R; writing—review and editing, Y.L., B.E., M.G., J.K, and A.R; supervision, J.K.; project administration, B.E.; funding acquisition, M.G., B.E., and A.R.

**Funding:** This research was funded by CPWR—The Center for Construction Research and Training—through cooperative agreement number U60-OH009762 from the National Institute of Occupational Safety and Health (NIOSH). Its contents are solely the responsibility of the authors and do not necessarily represent the official views of the CPWR or NIOSH.

**Data Availability Statement:** The data presented in this study are available on request from the corresponding author.

**Conflicts of Interest:** The authors declare no conflict of interest.